\documentclass[twoside]{article}
\usepackage{siunitx}
\usepackage{booktabs}
\usepackage[preprint]{aistats2027}
\usepackage[round]{natbib}

\usepackage{algorithmicx}
\usepackage{graphicx}
\usepackage{algcompatible}
\usepackage{algorithm}
\usepackage{algpseudocode}
\usepackage{xcolor}
\usepackage{amsmath}
\usepackage{amssymb}
\usepackage{subcaption} 
\usepackage{bm}
\usepackage{enumitem}
\DeclareMathOperator*{\argmax}{arg\,max}

\begin{document}

%

%

\twocolumn[

\aistatstitle{Example-driven Parametrisations for Bayesian Shape Optimisation}

\runningauthor{Diaz-Aylwin, Neighbour, Zhang, Talwar, Moss}

  \aistatsauthor{
    Gabriel Diaz-Aylwin  \And Joseph Neighbour  \And  Abiel Malkani Talwar
  }
  \aistatsaddress{
      Lancaster University \\ \texttt{g.diaz-aylwin@lancaster.ac.uk} \And Lancaster University \\ \texttt{j.neighbour@lancaster.ac.uk} \And Lancaster University \\ \texttt{a.m.talwar@lancaster.ac.uk}
  }
  \aistatsauthor{
      Rui-Yang Zhang \And Henry B. Moss 
  }
  \aistatsaddress{
      Lancaster University \\ \texttt{r.zhang26@lancaster.ac.uk} \And Lancaster University \\ \texttt{henry.moss@lancaster.ac.uk}
  }
]

\begin{abstract}
Bayesian optimisation is the natural tool for shape design when objectives are expensive and non-differentiable, but it needs a compact yet expressive parameterisation of the search space. Hand-crafting one is a complex endeavour requiring domain expertise, and often yields implicit infeasible regions, artificial bounds, and coupled, unordered coordinates. We instead learn the parameterisation from a collection of existing designs, applying principal component analysis to the deformations between shapes. The result is a linear, interpretable search space in which the number of components explicitly trades expressivity against dimensionality. Across aerofoils, wings, and radio-frequency cavities, spanning 2D geometry to 3D aerodynamics and electromagnetics, we show improved sample efficiency and the ability to explore beyond the confines of hand-crafted baselines.
\end{abstract}

\section{Introduction}

Bayesian optimisation (BO) is a principled, sample-efficient strategy for the
global optimisation of expensive, black-box objectives \citep{garnett2023bayesian}. By
maintaining a probabilistic surrogate of the objective and sequentially querying
points that maximise an acquisition function, BO routinely succeeds where
gradients are unavailable, and evaluations are limited, with applications
ranging from hyperparameter tuning \citep{snoek2012practical} to chemical synthesis
\citep{shields2021bayesian} and materials design \citep{frazier2015bayesian}.


\begin{figure*}[t]
\centering
\includegraphics[width=\textwidth]{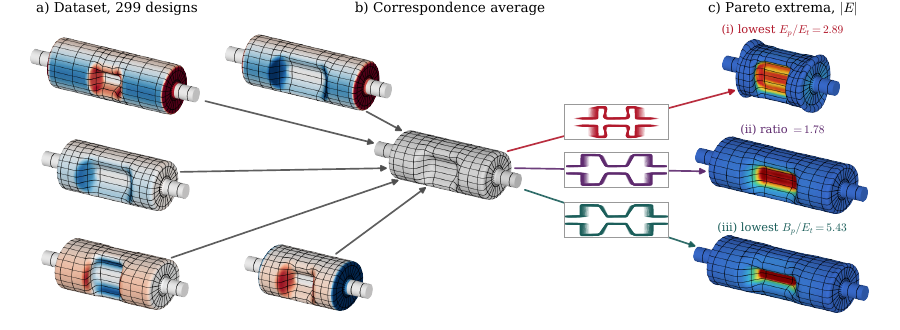}
\caption{BOLD's RF cavity designs, read left to right. 
\textbf{(a)} 
Some example cavities from our dataset, shaded by their normal displacement field from the correspondence mean (red outwards, blue inwards).
These, in essence, are the fields we run PCA on.
\textbf{(b)} 
The correspondence mean of the dataset (grey). We further draw $y = 0$ cross-sections along the straight line in shape coordinates interpolating from the mean to three designs on BOLD's Pareto front.
\textbf{(c)} The Pareto designs with their surface $|\mathbf{E}|$: the two ends of
the front, which minimise (i) $E_p/E_t$ (objective 1) and (iii) $B_p/E_t$ (objective 2), and (ii) the design
whose ratio $B_p/E_p$  matches a target specification. 
}
\label{fig:overview}
\end{figure*}

Shape design is a natural candidate for BO. Wherever geometry determines performance,
from aerofoils and wings to the radio-frequency (RF) cavities in particle accelerators, evaluating a candidate shape requires a costly simulation or physical
experiment, and gradients are frequently unavailable from black-box solvers.
Yet BO cannot be applied to shapes directly: it needs a compact Euclidean search
space, whereas a mesh describing a realistic geometry may contain
$10^5$--$10^6$ vertices. The success of BO for shapes therefore hinges almost
entirely on the choice of parametrisation.

This parametrisation choice typically falls to the practitioner. Expert parametrisations, such as
CAD feature parameters or free-form deformation
lattices, require substantial domain expertise and rarely
transfer between problems. They also make awkward search spaces: bounds must be
chosen artificially, large regions of the resulting box are infeasible or
self-intersecting, and coordinates cannot be ordered by importance.
Generic alternatives apply principal component analysis (PCA) \citep{jolliffe2002principal} to characteristic or signed distance functions
on a grid \citep{gaudrie2020modeling}, but linear combinations of these need
not produce valid shapes, requiring ad-hoc projection. Deep generative models require
large training sets \citep{durasov2021debosh, chen2025dora} and produce
nonlinear, uninterpretable latent spaces \citep{bodin2025linear}, while
gradient-based shape optimisation requires adjoint solvers and many iterations
\citep{wu2023framework}.

Designers, however, rarely start from scratch. A public aerofoil database,
previously optimised wings or past engineered geometries are usually
available, and a coherent family of shapes plausibly lies near a
low-dimensional region of shape space. We propose \emph{Bayesian Optimisation
by Local Deformation} (BOLD), which learns such a search space directly from
previous examples (Fig. \ref{fig:overview}a) using ideas from statistical shape analysis
\citep{dryden2016statistical}. 

BOLD represents each shape by its displacement
from a template and applies PCA to these displacements. The result is a
linear search space whose coordinates are ordered by explained variance, with
the number of components trading expressivity against dimensionality, and
linear no-fold constraints keep decoded shapes valid. We search it with a
TuRBO-style trust region \citep{eriksson2019scalable} centred on the best
design so far (Fig. \ref{fig:overview}b) and never clipped to a global box, so each step stays near a
shape that has already been evaluated, but over many iterations the search
can move far from the training data (Fig. \ref{fig:overview}c). In experiments on 2D aerofoils, 3D wings
and particle-accelerator cavities, BOLD is more sample-efficient than BO over hand-crafted
parametrisations and finds designs outside their support. On the RF-Dipole
cavity, to our knowledge the first application of BO to this design, our data-driven BOLD provides significant
improvements over a carefully hand-crafted baseline.

\section{Background}
\label{sec:background}

\subsection{Bayesian Optimisation}
Bayesian optimisation (BO) \citep{mockus1994application, garnett2023bayesian}
optimises expensive black-box objectives $f : \Theta \to \mathbb{R}$ over
$\Theta \subseteq \mathbb{R}^p$ by sequentially evaluating points that
maximise an acquisition function computed from a probabilistic surrogate.

\paragraph{Gaussian processes.} The surrogate is typically a Gaussian process
(GP) \citep{williams2006gaussian}. Conditioning on
$\mathcal{D}_n = \{(\theta_i, y_i)\}_{i=1}^n$, with
$y_i \sim \mathcal{N}(f(\theta_i), \sigma^2)$, GP renders a closed-form posterior
whose behaviour is determined by its kernel; e.g. Mat\'ern for Euclidean domains. 
Kernels can also be defined directly on meshes, such as
the Sliced Wasserstein Weisfeiler--Lehman (SWWL) kernel
\citep{perez2024gaussian}, which compares mesh embeddings.

\paragraph{Acquisition functions.} Given data $\mathcal{D}_n$, the next evaluation point is chosen as
$\theta_{n+1} = \argmax_{\theta \in \Theta} \alpha(\theta; \mathcal{D}_n)$ for some acquisition function $\alpha$. A
common $\alpha$ choice is the expected improvement (EI) \citep{jones1998efficient},
$\alpha_{\mathrm{EI}}(\theta; \mathcal{D}_n)
= \mathbb{E}_{f(\theta) \mid \mathcal{D}_n}[\max(f(\theta) - f^*, 0)]$, with
$f^* = \max_{i \le n} y_i$. EI has a closed form under a GP, but maximising it
is non-convex and typically requires multi-start gradient ascent over a bounded
$\Theta$. Effective search therefore needs $\Theta$ to be low-dimensional and
bounded, with every point corresponding to a valid design, which we will see in Section~\ref{sec:perils} is hard to achieve for shape optimisations.

\paragraph{BO extensions.} In high dimensions, trust-region methods such as
TuRBO \citep{eriksson2019scalable} and TREGO \citep{diouane2023trego} restrict
acquisition maximisation to a box around the incumbent that expands or
contracts depending on the recent query's evaluation.
Multi-objective BO (MOBO) \citep{daulton2022multi} targets improvements in the Pareto front, e.g.
through expected hypervolume improvement (EHVI) \citep{emmerich2006}.
Constrained BO models expensive black-box constraints with additional GPs and
weights the acquisition function by the probability of feasibility
\citep{gelbart2014bayesian}. Cheap constraints are imposed
directly during acquisition.

\subsection{Shape Representations}
\label{subsec:shapes}
Choosing how to represent shapes is the central problem of statistical shape
analysis \citep{dryden2016statistical}. We now review the three representation families that are most
relevant to shape optimisation.

\textbf{User-defined parametrisations.} These describe shapes by parameters
chosen by an expert, often taken from a CAD model or a domain-specific family.
For aerofoils, a common choice is the class-shape transformation (CST) of
\citet{kulfan2008universal}, which writes each surface as a fixed class
function, giving the round leading edge and sharp trailing edge, multiplied
by a weighted sum of Bernstein polynomials. They are compact and directly
optimisable, but are problem-specific and difficult to construct; thus they do not provide a path towards
general-purpose shape optimisation.

\paragraph{Implicit field representations.} A shape
$\Omega \subset \mathbb{R}^d$ can be encoded by its characteristic function (CF)
$\phi_{\mathrm{CF}}(x) = \mathbf{1}_{\{x \in\Omega\}}$, or by its signed distance function (SDF)
$\phi_{\mathrm{SDF}}$, equal to $-d(x, \partial\Omega)$ inside $\Omega$ and
$d(x, \partial\Omega)$ outside, where
$d(x, \partial\Omega) = \inf_{y \in \partial\Omega} \|x - y\|$. Both are
discretised on a fixed grid in practice. SDFs are smoother than CFs, but
neither is closed under linear combination: an average of two SDFs need not be
an SDF, nor describe a shape of the intended topology.

\paragraph{Deformation-based representations.} A third family represents each
shape by the transformation that maps a shared template onto it. Given pointwise
correspondences, the simplest choice is the displacement of each point from
the template, and doing PCA on these displacements gives the classical point
distribution model \citep{cootes1995active}. Diffeomorphic frameworks, such as
large deformation diffeomorphic metric mapping \citep{beg2005computing} and
stationary velocity fields (SVFs, \citet{arsigny2006log}), instead parametrise
deformations by smooth velocity fields with invertible flows, so deformed
shapes inherit the template's topology; PCA can then be applied to the
velocity fields \citep{vaillant2004statistics}. These parametrisations are well
established in medical imaging \citep{heimann2009statistical} but, to our
knowledge, have not been considered for BO.

\subsection{Bayesian Optimisation over Structures}
Most applications of BO to shape design optimise over a box in a user-defined
parametrisation \citep{chhantyal2025comparison, ghosh2021parametric,
kuszczak2023bayesian}. \citet{gaudrie2020modeling} instead apply PCA to CFs or
SDFs and run BO in the resulting eigenbasis. Others run BO in the latent space of a deep generative model
\citep{gomez2018automatic, moss2025return, fan2026search}; these need large
training sets, and their nonlinear latent spaces give distances no
direct geometric meaning and can produce invalid shapes
\citep{bodin2025linear}. BOLD instead learns a linear search space from modest
numbers of examples.

\subsection{Radio-Frequency Cavity Design}

The main application in this paper is the
RF-Dipole (RFD) design, a superconducting crabbing cavity used to increase collision
rates in particle colliders such as the Large Hadron Collider. Following
standard practice \citep{suba_thesis}, we minimise the peak surface electric and
magnetic fields, each normalised by the transverse field the cavity provides,
subject to the cavity resonating at \SI{394}{\mega\hertz}. RFD design is particularly challenging for BO:
existing BO methods for cavities optimise a 2D profile and sweep it into a
rotationally symmetric shape \citep{mobo_rf}, but the RFD has no rotational
symmetry; hand-crafted CAD parametrisations fail for some parameter
combinations, leaving infeasible regions in the search space and restricting
designs to what the CAD construction can express; implicit field
representations would need a fine 3D grid to resolve the features that govern
the peak fields, and offer no guarantee of valid shapes. Each
evaluation requires a 3D electromagnetic simulation; existing methods for RF cavity optimisation can take $\sim 50,000$ evaluations \citep{mobo_rf}, with full fidelity solves taking an hour of compute. To our knowledge, BO has not previously been applied to the RFD.


\section{The Perils of Shape Parametrisations}
\label{sec:perils}

Shape optimisation seeks the best shape within a family of designs sharing a common topology,
such as aerofoils, wings or cavities. 
We restrict to families within a common deformation class, that is, to shapes 
obtainable by smoothly
deforming a template shape $\Omega_0 \subset \mathbb{R}^d$.
Each design is parametrised by
$\theta \in \Theta \subset \mathbb{R}^p$, specifying a deformation $\phi_\theta$
of $\Omega_0$ which yields the shape $\Omega(\theta) = \phi_\theta(\Omega_0)$.
We thus wish to solve
\begin{equation*}
  \theta^* \in \operatorname*{arg\,max}_{\theta \in \Theta} f\bigl(\Omega(\theta)\bigr),
\end{equation*}
where $f$ is an expensive black-box objective function.

Data-driven approaches to this problem encode each shape as a
high-dimensional vector $e(\Omega) \in \mathbb{R}^D$, such as a discretised CF
or SDF, and reduce dimension with PCA \citep{gaudrie2020modeling}. Given
encodings of $N$ example shapes, with mean $\bar{e}$ and leading $p$
eigenpairs $(U_p, \Lambda_p)$ of their covariance, each shape is represented
by its whitened coordinates
$r(\Omega) = \Lambda_p^{-1/2} U_p^\top (e(\Omega) - \bar{e}) \in \mathbb{R}^p$.
Whitening gives the examples unit variance in every direction, so a unit step
means one standard deviation of variation along any coordinate.
BO is run over a box $[-c,c]^p$, and each candidate $z$ is decoded to
$d(z) = \bar e + U_p\Lambda_p^{1/2}z$ and from there to a shape. 

In the remainder of this section,
we identify two pathological issues with this pipeline that we argue are the fundamental obstacles for effective BO over shapes.


\begin{figure}
    \centering
    \captionsetup{font=small, skip=3pt}
    \captionsetup[subfigure]{skip=1pt}

    \begin{subfigure}[b]{0.48\columnwidth}
        \centering
        \includegraphics[width=\textwidth, trim=6 6 6 6, clip]{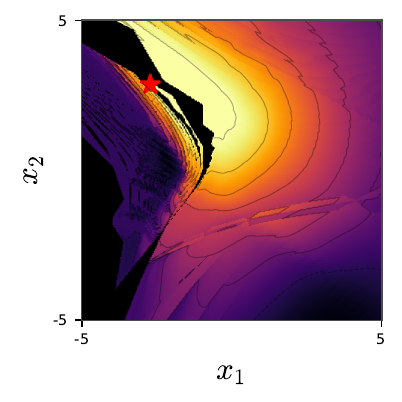}
        \caption{Signed Distance Func.}
        \label{fig:obj_land_sdf}
    \end{subfigure}\hfill
    \begin{subfigure}[b]{0.48\columnwidth}
        \centering
        \includegraphics[width=\textwidth, trim=6 6 6 6, clip]{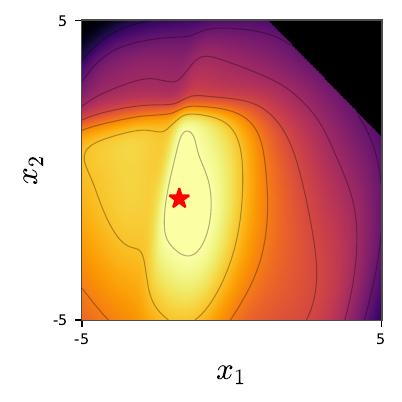}
        \caption{CST}
        \label{fig:obj_land_cst}
    \end{subfigure}

    \vspace{2pt}

    \begin{subfigure}[b]{0.48\columnwidth}
        \centering
        \includegraphics[width=\textwidth, trim=6 6 6 6, clip]{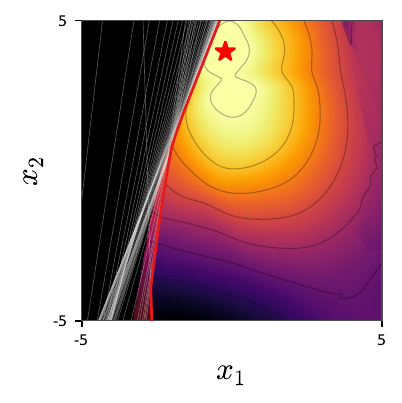}
        \caption{DF + no-fold}
        \label{fig:obj_land_dp_nofold}
    \end{subfigure}\hfill
    \begin{subfigure}[b]{0.48\columnwidth}
        \centering
        \includegraphics[width=\textwidth, trim=6 6 6 6, clip]{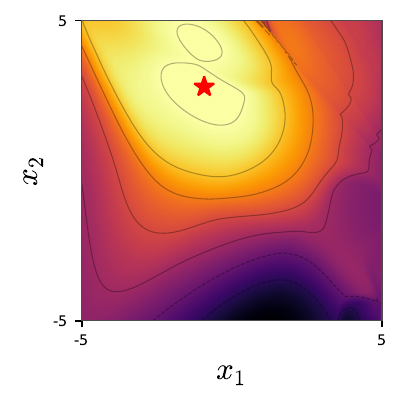}
        \caption{Vector Field}
        \label{fig:obj_land_vf}
    \end{subfigure}
    
    \vspace{2pt}
    
    \caption{We plot an objective function landscape (lift-to-drag ratio) across the leading components for a variety of shape representations. In black, we denote the region in which decoding linear combinations of the leading modes yields an invalid aerofoil. Note that the landscape is jagged for the SDFs (a), has clear infeasible regions (black) for the expert-designed CST parametrisation (b) and DFs (c), but is well-behaved for the SVFs (d), since these preserve injectivity by construction. In (c), we further plot DF with our no-fold constraints (white lines), which approximate the infeasible region by linear inequalities and the DFs' infeasible boundary (red).}
    \label{fig:landscape}
\end{figure}

\subsection{Representations Can Generate Invalid Shapes and Non-smooth Objectives}
\label{sec:invalid}

Decoding a point in the search space returns a new encoding, but these encodings are not guaranteed to yield a valid shape. The average of two SDFs is not an SDF, for instance, and its zero
level set may be disconnected. Hand-crafted parametrisations have the same
issue when the CAD model self-intersects or fails to build
(Figure~\ref{fig:landscape}). 

Implicit representations also
make the objective harder to model. Drag and similar objectives usually
change smoothly with geometry, but in an implicit representation $z$ controls
a field, and the shape is defined only by that field's zero level set. For an
aerofoil, a small move
in $z$ can remove a trailing edge and cause a sharp change in
drag that a GP will struggle to model. 

\textit{BOLD instead represents shapes
as deformations of the current best shape, so small steps in $z$ give small
changes in shape, and imposes linear constraints so that every feasible point
decodes to a valid shape.}

\subsection{Search Spaces Need Bounds}
\label{sec:bounds}

Acquisition optimisation needs a bounded domain. For hand-crafted
parametrisations, bounds are set by expert judgement and directly limit the
designs that can be found. Whitened PCA coordinates seem to offer a principled
choice, but the box size $c$ remains a problem. The whitened examples have mean
squared norm $p$, while a point drawn uniformly from $[-c, c]^p$ has expected
squared norm $pc^2/3$. Any box wide enough to contain most of the examples
(say $c = 3$) therefore has most of its volume well outside, where the
learnt representation has no reason to produce sensible geometry, while a
smaller box excludes potentially good designs. In practice, results are sensitive to this choice
(Appendix~\ref{app:boxsize}). 

\textit{BOLD avoids fixing a global box. Instead, we focus on an adaptively-sized box around the best shape found so far: the search stays close to shapes the representation describes
well, while remaining free to move beyond the examples over successive
iterations.}

\section{BO by local deformations}
\label{sec:BOLDDMM}

We now present our main contribution, Bayesian Optimisation by Local
Deformation (BOLD). BOLD represents each shape by its displacement from a
template shape, applies PCA to these displacements to get a low-dimensional
linear search space, and adds linear no-fold constraints to keep decoded
shapes valid (Section~\ref{sec:deformations}). We search this space with a
trust region centred on the best shape found so far, which removes the need
for global bounds (Section~\ref{sec:recentring}).
See Algorithm~\ref{alg:bold} for a high-level overview of BOLD. Our proposed data-driven parametrisation relies on
correspondences between shapes, and Section~\ref{sec:correspondence} describes
their construction for the shapes studied in
Section~\ref{sec:experiments}.

\begin{algorithm}[t]
\caption{BOLD: Bayesian Optimisation by Local Deformation (examples $\{\Omega_i\}_{i=1}^N$, initial data
         $\mathcal{D}_0$, budget $T$,
         initial trust-region size $L$)}
\label{alg:bold}
\begin{algorithmic}[1]
\For{$n = 0, \dots, T-1$}
  \State $\Omega_0 \gets \argmax_{(\Omega, y) \in \mathcal{D}_n} y$
  \State Re-centre displacement field around $\Omega_0$
  \State Fit GP to the encoded data
  \State $z_{n+1} \gets \arg\max_{z \in [-L, L]^p,\; Az \geq b}
    \alpha_{\mathrm{EI}}(z; \mathcal{D}_n)$
  \State $\Omega_{n+1} \gets \Omega(z_{n+1})$
  \State $y_{n+1} \gets f(\Omega_{n+1}) + \epsilon_{n+1}$
  \State $\mathcal{D}_{n+1} \gets \mathcal{D}_n \cup \{(\Omega_{n+1}, y_{n+1})\}$
  \State Update trust-region size $L$
\EndFor
\State \Return best design in $\mathcal D_T$
\end{algorithmic}
\end{algorithm}

\subsection{A Search Space of Deformations}\label{sec:deformations}
 
Ideally, every point in the search space would decode to a valid shape, and
nearby points would decode to similar shapes. CFs and SDFs manage neither
(Section~\ref{sec:invalid}). Deforming a valid template is one way to get
valid shapes, because an injective deformation keeps the template's
topology\footnote{This rules out topology
optimisation~\citep{bendsoe2013topology}, which is outside the scope of this
paper.}. The diffeomorphic methods of Section~\ref{subsec:shapes} are
injective by construction (Figure~\ref{fig:airfoil_registration_eg}a), but
each shape's velocity field has to be found by non-convex optimisation, and
this must be repeated whenever the template changes and becomes too expensive for detailed 3D geometries such as RF cavities. We
therefore work with displacement fields, which are cheap to compute, trivial to recentre, and handle
validity separately through linear constraints.


\begin{figure}[t]
  \centering
  \begin{subfigure}[b]{0.48\columnwidth}
    \centering
    \includegraphics[width=\textwidth]{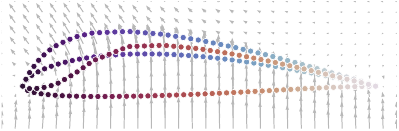}
    \caption{Stationary velocity field}
  \end{subfigure}
  \begin{subfigure}[b]{0.48\columnwidth}
    \centering
    \includegraphics[width=\textwidth]{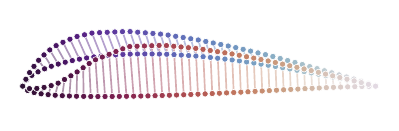}
    \caption{Displacement field}
  \end{subfigure}
\caption{Two representations of the deformation between a pair of aerofoils.
Points of the same colour correspond. (a) A stationary velocity field (grey
arrows), defined over the whole plane, whose flow carries one aerofoil onto
the other. (b) A displacement field, which moves each point directly to its
corresponding point (lines).}
  \label{fig:airfoil_reps}
\end{figure}

\paragraph{Displacement fields.}
Suppose each shape is represented by $P$ points $X \in \mathbb{R}^{P \times d}$
in \emph{correspondence}, so that the $j$-th point marks the same geometric
feature, such as the leading edge of an aerofoil, on every shape. We encode a
shape by its displacement from a template shape $X_*$, $e(X) = \operatorname{vec}(X - X_*)$,
which moves every point directly to its counterpart
(Figure~\ref{fig:airfoil_reps}b), and then apply the PCA pipeline of
Section~\ref{sec:perils} with the template in place of the mean $\bar{e}$.
Decoding is affine, $\operatorname{vec} X(z) = \operatorname{vec} X_* + U_p \Lambda_p^{1/2} z$. Coordinates are ordered by
explained variance, and $p$ trades expressivity against dimensionality. Note that updating $X_*$ corresponds to a simple translation in the encoded space.


\paragraph{No-fold constraints.}
Unlike SVFs, linear combinations of displacement fields can result in a folded shape.  We prevent
this locally by requiring that each edge $e$ of the correspondence grid keeps
its orientation on the template, $\Delta_e(0) \cdot \Delta_e(z) \ge \delta$,
where $\Delta_e(z) = X_k(z) - X_j(z)$ is the edge vector of the decoded shape and $\delta > 0$ is
a margin. Since decoding is affine, these reduce to linear inequalities
$Az \ge b$, imposed directly during acquisition optimisation
(Figure~\ref{fig:landscape}c). The margin \(\delta\) (initialised at $0.2mm$) can be increased adaptively in the loop if we are getting mesh failures despite the constraints holding.
For example, adding no-fold conditions to the
edges between corresponding upper and lower surface points stop aerofoil
surfaces from crossing, and to edges along and between rings stop RF cavity grid
lines from reversing order.



\subsection{Bound-free Search by Re-centring}\label{sec:recentring}
 We want the search to explore freely, including well beyond the examples,
while only ever proposing shapes the representation describes well. A fixed
box cannot do both (Section~\ref{sec:bounds}), so we instead build on TuRBO~\citep{eriksson2019scalable} and
take the incumbent (the best evaluated design so far) as the template $X_*$. The
trust region is then simply $[-L, L]^p$ around $z = 0$, and $L$ is doubled
after $3$ consecutive improvements and halved after
$5$ consecutive failures
. 
How far the search should move from the examples
depends on the problem, so clipping the trust region to a global box is optional. A global box keeps proposals near the data, which suits objectives that are unreliable away from it, such as neural-network proxy objective functions (Sec. \ref{sec:eg-aerofoils},\ref{sec:eg-wings}). Leaving the trust region unclipped lets the search travel arbitrarily far, which we found to suit trusted numerical simulators, such as in our RF cavity design (Sec. \ref{sec:eg-cavities}).


\subsection{Constructing Correspondences}
\label{sec:correspondence}
In general, correspondence is an ill-defined problem: two surfaces with the same topology admit infinitely many
bijections. For PCA to be
meaningful, the correspondence should match genuinely equivalent features, so
that displacement fields reflect differences in geometry rather than in
sampling. In practice, many shape design tasks have structure that makes this
straightforward: landmarks such as leading and trailing edges, a known
topology, or a natural direction along the shape. We give two examples.

\begin{figure}[t]
  \centering
  \begin{subfigure}[b]{0.48\columnwidth}
    \centering
    \includegraphics[width=\columnwidth]{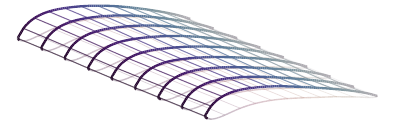}
    \caption{Wing}
    \label{fig:wing}
  \end{subfigure}
  \begin{subfigure}[b]{0.48\columnwidth}
    \centering
    \includegraphics[width=\textwidth]{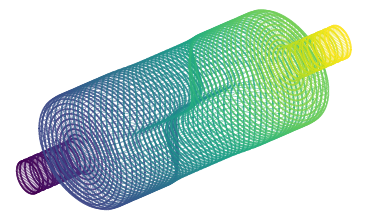}
    \caption{RF cavity}
    \label{fig:rf}
  \end{subfigure}
  \caption{Structured grids used to build correspondences. (a) A wing sampled
  at common spanwise sections, each sampled uniformly in arc length. (b) An RF
  cavity gridded by level sets of the harmonic coordinate $h$ (rings, coloured
  by $h$) and arc length along each ring. In both cases, points with the same
  grid index correspond across all shapes in the dataset.}
  \label{fig:correspondence_examples}
\end{figure}

\paragraph{Aerofoils and wings.} We locate the leading and trailing edges of
each aerofoil \citep{seliguiuc}, split the contour into upper and lower
surfaces, and sample each surface uniformly in arc length with a fixed number
of points. Wings \citep{diniz2025optiwing3d} are sampled at a common set of
spanwise locations, and each section is treated as an aerofoil
(Figure~\ref{fig:wing}). 

\paragraph{RF cavities.} A cavity wall is topologically a tube between the two
beam-pipe openings. For a longitudinal coordinate, we compute the harmonic
function $h : S \to [0, 1]$ equal to $0$ at one opening and $1$ at the other,
which minimises the Dirichlet energy $\int_S \|\nabla_S h\|^2 \,\mathrm{d}A$
\citep{floater2005surface} and so solves the Laplace--Beltrami equation
$\Delta_S h = 0$ \citep{pinkall1993computing}; discretising $\Delta_S$ with the
cotangent Laplacian gives a sparse linear system. The level sets of $h$ are
smooth rings around the cavity, along which we place points uniformly in arc
length. To align prominent features such as the ends of the central barrel
across the dataset, we fix the values of $h$ at which they occur and sample
uniformly between them, giving each segment a number of rings proportional to
its mean length over the dataset (Figure~\ref{fig:rf}). 

\section{EXPERIMENTAL RESULTS}
\label{sec:experiments}
We evaluate BOLD on three problems of increasing difficulty: 2D aerofoils, 3D
wings and 3D RF cavities. On the aerofoils and wings, we compare BOLD using displacement
fields (DP-PCA) and SVFs (VF-PCA) against hand-crafted CST parametrisations of
the same dimension, and on the aerofoils also against PCA on SDFs
(following \citet{gaudrie2020modeling}). On the RF cavity, we compare
against BO over the CAD parameters that generated the dataset, as it was computationally prohibitive to calculate accurate SVFs for the fine meshes required for reliable  evaluations. Unless stated
otherwise, the surrogate is a GP with an ARD Mat\'ern-5/2 kernel on each
method's coordinates. For 3D problems, when computational resources allowed, we also used SWWL
kernels \citep{perez2024gaussian}, which compare the decoded meshes directly
and so remove any dependence of the surrogate on the parametrisation choice. Within each
problem, all methods share the same initial designs, evaluation budget,acquisition function and trust region logic, and we report the best objective found
against the number of evaluations over $30$ seeds. Full experimental
details are in Appendix~\ref{app:experiments}, with ablations on global
box bounds and number of PCA components in Appendices~\ref{app:boxsize}
and~\ref{app:PCA}.

\subsection{2D Aerofoils} \label{sec:eg-aerofoils}

\begin{figure*}[tp]
    \centering
    \includegraphics[width=\textwidth]{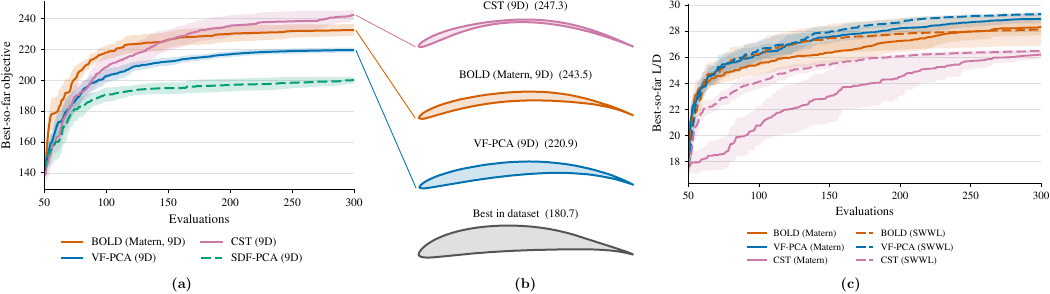}
    \caption{Aerodynamic shape optimisation.\textbf{(a)} For 2D aerofoils, BOLD recovers most of the performance of the best hand-tuned parametrisation.
\textbf{(b)}~  BOLD comes close to the best CST airfoil without needing a hand-crafted design. 
\textbf{(c)}~ 3D wing optimisation: VF-PCA and BOLD (first 15 modes) and a hand-designed sectional CST parameterisation. BOLD and VF-PCA are statistically indistinguishable and outperform CST, which is also sensitive to the choice of Mat\'ern kernel or SWWL surrogate. BOLD shows little dependence on the kernel.}
    \label{fig:airfoils_wings}
\end{figure*}

Our first experiment (Figure~\ref{fig:airfoils_wings}a) tests whether BOLD,
learning its search space purely from data, can match a strong hand-crafted
parametrisation. We use aerofoil design, where the Kulfan (CST)
parametrisation \citep{kulfan2008universal} is the established choice. We build data-driven representations using the UIUC coordinate database \citep{seliguiuc} of 1657
airfoils. Our objective is the lift-to-drag ratio at $\alpha = 4^\circ$,
$Re = 10^6$. Since we rely on this example to investigate a large number of ablations, 
we compute this with  NeuralFoil \citep{sharpe2025neuralfoil}, a neural-network proxy for lift-to-drag. 
To account for the objective's limitations, we multiply the objective by the surrogate's confidence.

\subsection{3D Wings} \label{sec:eg-wings}
Our second experiment (Figure~\ref{fig:airfoils_wings}c) moves to 3D wing
design, a higher-dimensional problem where hand-crafted parametrisations
become less flexible, and are comfortably outperformed by BOLD. We initialise data-driven representations using the OptiWing3D dataset \citep{diniz2025optiwing3d} containing 776 wings, each the result of
a lift-constrained drag minimisation. 
We then seek to maximise the lift-to-drag ratio of the full wing at Mach 0.59,
$Re = 5.6\times 10^6$ (the dataset medians) and $\alpha = 4^\circ$. We compute this with AeroBuildup, a module of the AeroSandbox package \citep{aerosandbox_phd_thesis, aerosandbox},
which couples NeuralFoil \citep{sharpe2025neuralfoil} with a lifting-line induced-drag model. 
We compare against the hand-crafted parameterisation of a sectional Kulfan parameterisation using $3$ 2d parameterisation of aerofoils at root, mid and tip sections with interpolation between ($3\times5$ parameters).



\subsection{RF Cavities} \label{sec:eg-cavities}


\begin{figure}[t] 
\centering
\begin{subfigure}[t]{0.49\columnwidth}\centering\includegraphics[width=\linewidth]{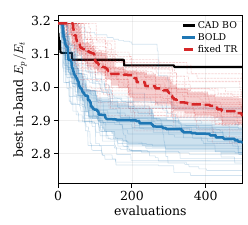}\caption{}\label{fig:cav-so}\end{subfigure}\hfill
\begin{subfigure}[t]{0.49\columnwidth}\centering\includegraphics[width=\linewidth]{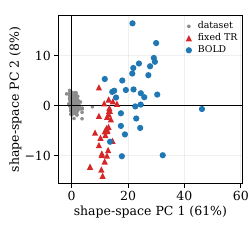}\caption{}\label{fig:cav-pca}\end{subfigure}\\[2pt]
\begin{subfigure}[t]{0.49\columnwidth}\centering\includegraphics[width=\linewidth]{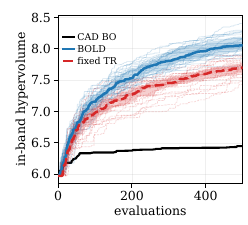}\caption{}\label{fig:cav-hv}\end{subfigure}\hfill
\begin{subfigure}[t]{0.49\columnwidth}\centering\includegraphics[width=\linewidth]{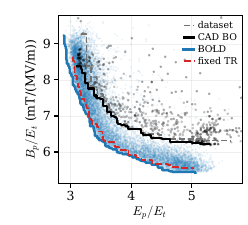}\caption{}\label{fig:cav-pareto}\end{subfigure}
\caption{RF cavity optimisation. \textbf{(a)} Best in-band $E_p/E_t$
(lower is better) and \textbf{(c)} in-band hypervolume against evaluations,
showing medians, quartiles and the 30 seeds. BOLD outperforms CAD BO
in every seed. \textbf{(b)} The dataset and the single-objective
optima projected onto their two leading principal components. BOLD is able to explore far beyond the initial shapes. 
\textbf{(d)} Pooled
in-band Pareto fronts, with the dataset for reference. The front found by BOLD using EHVI dominates both baselines.
}
\label{fig:cavities}
\end{figure} 

\begin{figure}[h!]
  \centering
  \includegraphics[width=\columnwidth]{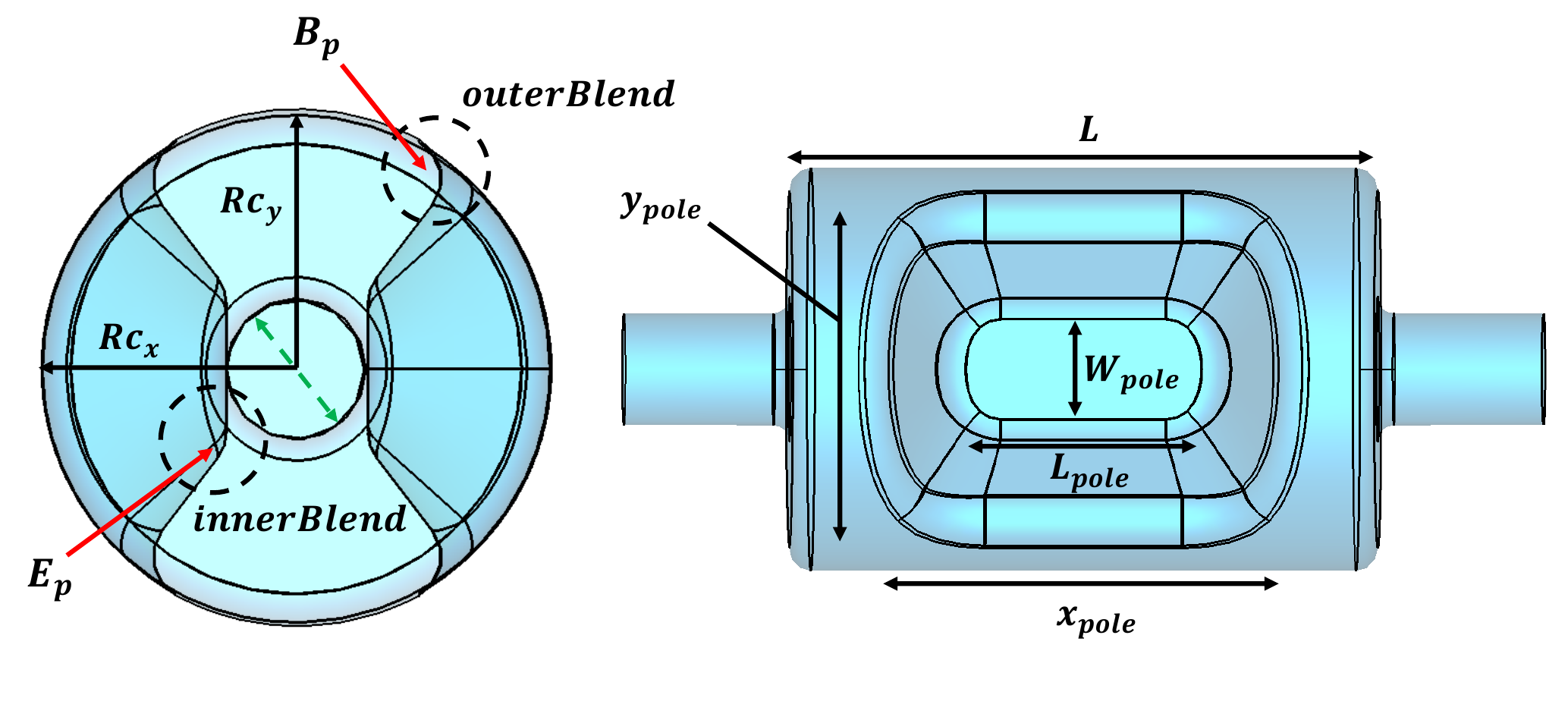}
  \caption{The nine-parameter CAD model of the RF-dipole cavity used to generate the initial shape dataset and BO baselines. Cavity length $L$, the pole width $W_\mathrm{pole}$ and length
$L_\mathrm{pole}$, the transverse radii $Rc_{x}$ and $Rc_y$, and the inner and outer
blend radii are labelled. Red arrows mark where the peak surface fields occur: $E_p$ on the
pole tips and $B_p$ at the pole--body junction. The green dashed line marks
the 100~mm aperture, which must be preserved by all designs.}
  \label{fig:cad_par}
\end{figure}

Our final experiment is the design of a 3D RF-dipole cavity, considering both single- and multi-objective shape design. We build the
BOLD search space from 299 cavities generated from a carefully hand-designed
nine-parameter CAD model (Figure~\ref{fig:cad_par}), evaluated in the proprietary solver CST
Microwave Studio \cite{CST} on an 8-core CPU with 32~GB of RAM. To avoid confusion with the Class Shape Transformation introduced in Section \ref{sec:eg-aerofoils}, this solver is hereafter referred to as the CAD solver.
This problem is too expensive for the other representations in our study.
Registering SVFs to every cavity is impractical for detailed 3D geometry, and
SDFs would need a fine 3D grid to resolve the features that control the peak
fields. We therefore compare
BOLD only with standard BO over the CAD parameters. For the multi-objective runs, we replace EI with expected hypervolume
improvement (EHVI) \citep{emmerich2006}. Since there is no single
incumbent, we centre trust regions on the Pareto-optimal design with the largest hypervolume contribution, i.e. the one whose removal maximally reduces hypervolume.

For each design, we solve the source-free, lossless Maxwell eigenproblem for
the electric field $\mathbf{E}(\mathbf{x})\,e^{i\omega t}$ in the vacuum
$\Omega$ enclosed by the cavity wall $\Gamma$,
\begin{equation*}
\nabla\times\left(\mu_0^{-1}\,\nabla\times\mathbf{E}\right) = \omega^2\varepsilon_0\,\mathbf{E}
\quad\text{in }\Omega,
\quad
\mathbf{n}\times\mathbf{E} = 0 \quad\text{on }\Gamma,
\end{equation*}
treating the wall as a perfect conductor, as in superconducting operation.
From the solution we extract the peak surface fields
$E_p = \max_{\Gamma}|\mathbf{E}|$, $B_p = \max_{\Gamma}|\mathbf{B}|$, and the deflecting field $E_t$. We minimise $E_p/E_t$ and $B_p/E_t$ subject to
the deflecting mode lying within $394 \pm 1$~MHz. 
For both methods, this frequency is modelled by a GP and the acquisition function is multiplied by the resulting probability of lying in band.

Even this carefully designed CAD space contains parameter combinations that
fail to mesh, so the baseline needs an extra GP classifier, whose probability
of feasibility multiplies the acquisition function. BOLD needs no such classifier,
because its no-fold constraints keep proposals valid. The baseline must use
the CAD solver, and licensing limits how many runs we can
do. BOLD needs no CAD model, so we evaluate its designs with the open-source
finite-element solver Palace \citep{palace2023} (with performance verified in Fig. \ref{fig:validate}).

\paragraph{Results.} Figure~\ref{fig:cavities} compares BOLD with constrained BO over the nine CAD
parameters (CAD BO), both starting from the same 299 cavities. To isolate the
effect of leaving the initial data, we also run BOLD with its trust region clipped to
a global box of $\pm 6\sigma$, the smallest that contains the dataset. BOLD clearly outperforms CAD BO: even its worst of 30 seeds
beats the CAD run on both the single- and multi-objective problems. The
clipped variant recovers much of this gain, so most of the improvement comes
from the learned shape space, but removing the box gives a further, consistent
improvement, exploiting BOLD's ability to leave the distribution of the
generating parametrisation while still proposing valid designs.
Figure~\ref{fig:cavities}b projects the dataset and the single-objective
optima onto their leading principal components. The dataset forms a tight
cluster, while the optima lie far outside it, with the unclipped optima the
most extreme. Every design on the pooled Pareto front has at least one
coordinate outside the range of the dataset (Figure~\ref{fig:box-and-whisker}).

\begin{figure}[h]
\centering
\includegraphics[width=0.75\columnwidth]{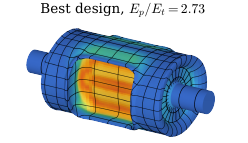}\\[-2pt]
\includegraphics[width=0.49\columnwidth]{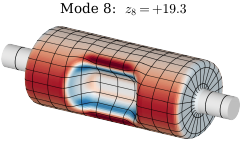}\hfill
\includegraphics[width=0.49\columnwidth] {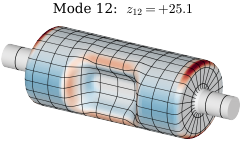}
\caption{The lowest-$E_p/E_t$ design of the single-objective sweep ($E_p/E_t$ 2.73, $B_p/E_t$ 13.65). 
As in Figure \ref{fig:overview}, we overlay its $| \textbf{E}|$ field. Below, we add its two most outlying shape coordinates, 
shaded by their normal displacement field from the correspondence mean 
(red outwards, blue inwards).
}
\label{fig:best_so}
\end{figure}

\paragraph{Discovering new important geometry.}
BOLD's optima are physically sensible (see Figure~\ref{fig:overview}(c)) and several rely on
features that our hand-crafted parametrisation could not express.
$E_p$ peaks on the pole tips and $B_p$ at the pole--body junction, so reducing them calls for smooth curvature and more
surface area in these regions, while the frequency constraint limits the
transverse size of the cavity. The cavity shown in Figure \ref{fig:overview}(c)(i) has a wider pole in $y$, which spreads the field over
a larger surface and also increases $E_t$. Its pole tips bulge outwards,
adding surface area without reducing the aperture, a shape the CAD model
cannot produce. Its end-caps are lobed,
which adds volume near the ends and lets the frequency be tuned without
changing the poles. This design improves $E_p/E_t$ by 13.8\% over the best
baseline. A design that trades off both objectives (Figure~\ref{fig:overview}(c)(ii))
improves mainly through smoother pole shapes and larger cavity length. Single objective optimisations were free to explore more extreme regions of the shape space; the cavity shown in Figure \ref{fig:best_so} features a re-entrant end-cap and a pronounced pole--can junction which increases $E_t$ and the surface area of the poles, respectively.  BOLD has identified key geometric changes, which will guide future
CAD parametrisations of RF cavities.

\section{Discussion}
We have shown that a search space for BO can be learned from a modest number
of existing designs, by applying PCA to the displacements between shapes in
correspondence. The resulting space is linear and low-dimensional, and simple
linear constraints can help keep the decoded shapes valid. We search it with a trust
region around the best design so far, and the user can choose whether to clip
this to a global box depending on how far they want to move from the initial data.
BOLD matched the custom CST parametrisation for aerofoils and beat it for wings. For
the RF-dipole cavity, it beat BO over a hand-designed CAD model, did not need
a classifier to avoid failed designs, and produced features the CAD model
cannot represent.

There are several limitations. The topology is fixed by the template, and
the search can only leave the dataset along modes that appear in the
examples. Correspondences have to be built for each shape family, although
this was simple for the families we considered. Our no-fold constraints only
act on neighbouring points, so they cannot rule out global self-intersection,
although this did not occur in our experiments. Constraints that guarantee
injectivity, and ways of choosing the global box automatically, are left for
future work.


\bibliography{references}

\newpage
\appendix
\onecolumn
\aistatstitle{Appendix}

\section{Shape Deformations via Stationary Vector Fields}
\label{app:SVF}

This appendix gives the details of the SVF registration used in
Section~\ref{sec:deformations}. Computing an SVF requires three choices: a
finite-dimensional parametrisation of the velocity field, a data term measuring
agreement between the deformed template and the target, and a regulariser.

\subsection{Velocity field parametrisation}

We represent the velocity field by tensor-product cubic B-splines on a regular
control lattice covering a domain $\mathcal{X} \subset \mathbb{R}^d$ that
contains all shapes. In two dimensions,
\[
  v(x, y) = \sum_{m,n=1}^{S} c_{m,n} \, B_m(x) \, B_n(y),
  \qquad c_{m,n} \in \mathbb{R}^2,
\]
where $B_m(x) = B((x - x_m)/\Delta x)$ is the cubic cardinal B-spline centred at
control point $x_m$ with spacing $\Delta x$, and
\[
  B(x) =
  \begin{cases}
    2/3 - |x|^2 + |x|^3/2 & |x| < 1, \\
    (2 - |x|)^3/6 & 1 \le |x| < 2, \\
    0 & \text{otherwise},
  \end{cases}
\]
and likewise for $B_n(y)$. The extension to $d$ dimensions is immediate, giving
$d S^d$ coefficients in total, which we collect into a vector $c$. Each basis
function has compact support, so the velocity at any point depends on only
$4^d$ coefficients.

\subsection{Registration objective}

Given a template represented by points $\{x_0^i\}_{i=1}^P$ and a target
$\Omega_t$, the deformed template is obtained by integrating
\[
  \frac{\mathrm{d}}{\mathrm{d}t}\varphi_t(x) = v_c\big(\varphi_t(x)\big),
  \qquad \varphi_0(x) = x,
\]
over $t \in [0, 1]$ with $Q$ steps of a fourth-order Runge--Kutta scheme, and
evaluating $\varphi_1(x_0^i)$. We minimise
\[
  \mathcal{L}(c) = D\big(\{\varphi_1(x_0^i)\}_i, \Omega_t\big)
  + \lambda \, \|v_c\|_{\mathrm{Sob}}^2 .
\]

\paragraph{Data term.} When a pointwise correspondence $x_0^i \leftrightarrow
x_t^i$ is available, we use the sum of squared distances between corresponding
points,
$D = \sum_{i} \|\varphi_1(x_0^i) - x_t^i\|_2^2$. Otherwise, we sample both
boundaries approximately uniformly and use the symmetric Chamfer distance,
\[
  d_C(A, B) = \sum_{a \in A} \min_{b \in B} \|a - b\|_2^2
  + \sum_{b \in B} \min_{a \in A} \|a - b\|_2^2 .
\]

\paragraph{Regulariser.} The inverse problem is not unique, so we favour
smooth velocity fields by penalising the discrete Sobolev semi-norm of the
coefficients,
\[
  \|v_c\|_{\mathrm{Sob}}^2 = \sum_{m=1}^{S-1} \sum_{n=1}^{S}
  \|c_{m+1,n} - c_{m,n}\|_2^2
  + \sum_{m=1}^{S} \sum_{n=1}^{S-1} \|c_{m,n+1} - c_{m,n}\|_2^2 ,
\]
a finite-difference approximation to the Dirichlet energy
$\int_{\mathcal{X}} \|\nabla v(x)\|_F^2 \,\mathrm{d}x$.
\subsection{Optimisation}

Optimising directly on a fine lattice is difficult, since the objective is
highly non-convex and the field has many degrees of freedom. We therefore use a
coarse-to-fine schedule. Optimisation begins on a coarse lattice with few
boundary samples and a large $\lambda$, which favours smooth, global
deformations. The resulting field is projected onto a finer B-spline basis and
used to initialise the next stage, which uses a denser lattice, more boundary
samples and a smaller $\lambda$. Early stages thus capture large-scale
differences between the shapes, and later stages refine local detail. At each
stage we run Adam followed by L-BFGS. In our experience, this gives more stable
optimisation and tighter registrations than optimising at the finest
resolution directly.

\subsection{Computational cost}

Each evaluation of $\mathcal{L}$ integrates the flow for all $P$ points, which
requires $4Q$ velocity evaluations per point. With compactly supported
B-splines, each velocity evaluation costs $\mathcal{O}(4^d d)$, so the flow
costs $\mathcal{O}(Q P 4^d d)$. The Chamfer distance costs
$\mathcal{O}(P \log P)$ with KD-trees, or $\mathcal{O}(P^2)$ by brute force,
and the regulariser $\mathcal{O}(d S^d)$. Gradients by automatic
differentiation add a constant factor. The total cost of a registration is
this per-evaluation cost multiplied by the number of Adam iterations and
L-BFGS function evaluations, summed over stages.

\subsection{Aerofoil example}

\begin{figure}[t]
  \centering
  \includegraphics[width=\linewidth]{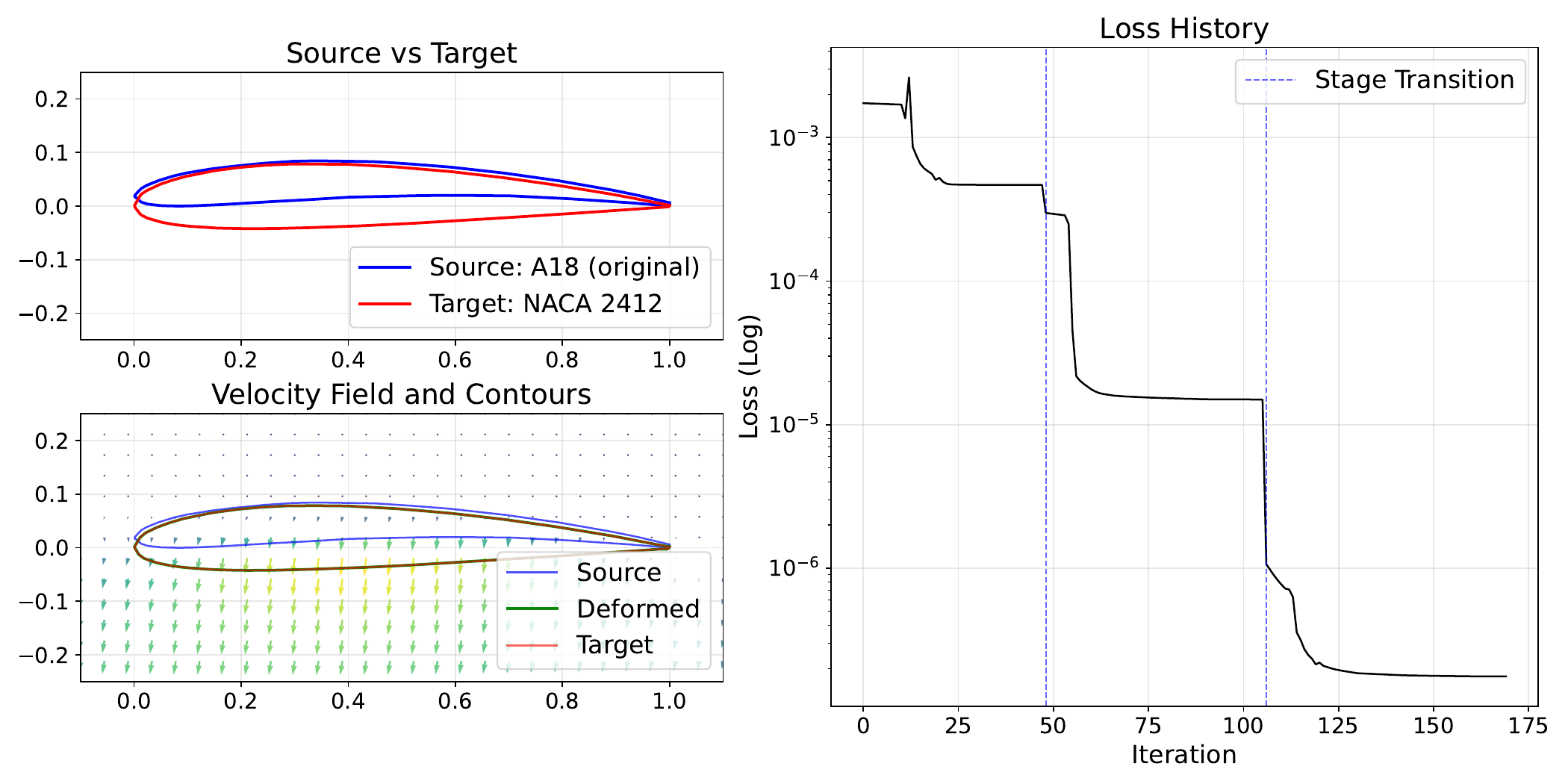}
  \caption{SVF registration between two aerofoils. Top left: source and
  target after chord normalisation and arc-length resampling. Bottom left: the
  learnt velocity field with the source, target and deformed source. Right:
  the loss on a log scale, with dashed lines marking transitions between
  coarse-to-fine stages.}
  \label{fig:airfoil_registration_eg}
\end{figure}

\begin{figure}[t]
  \centering
  \includegraphics[width=0.8\linewidth]{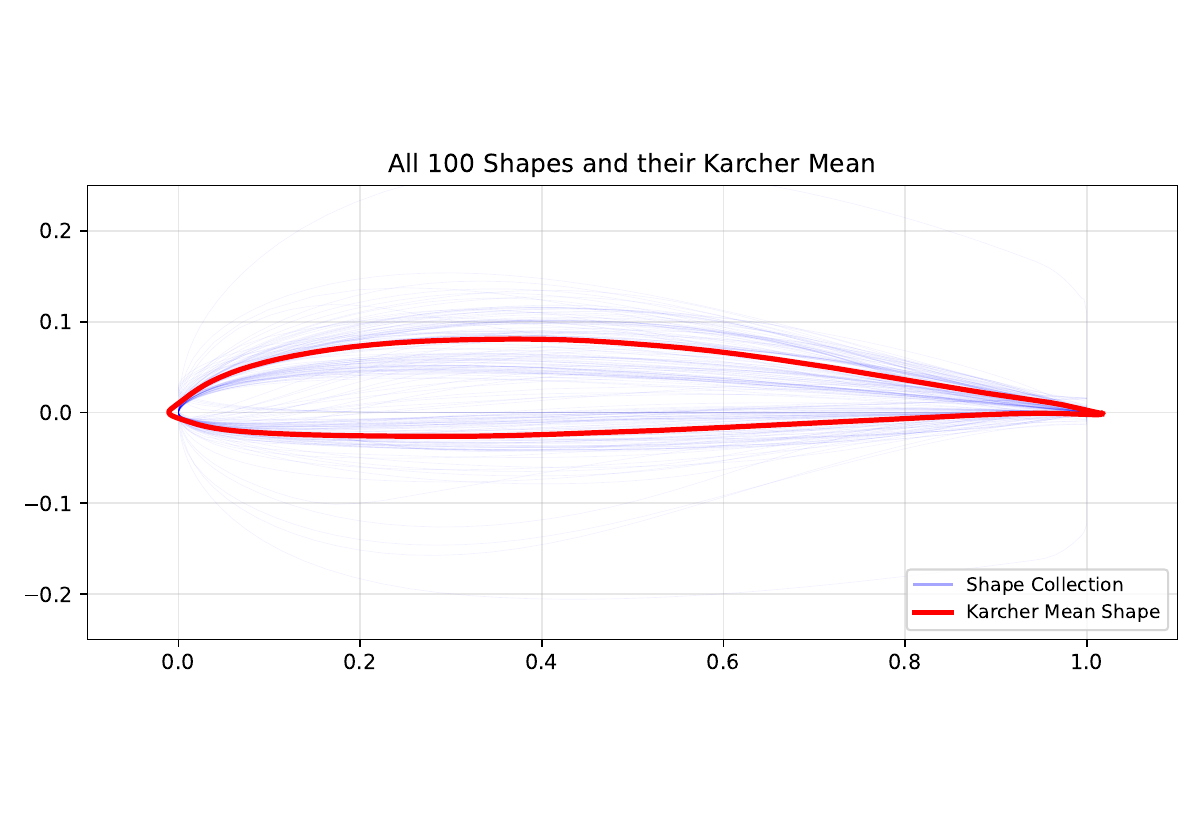}
  \caption{Mean shape of 100 normalised aerofoils (red), computed by
  registering the pointwise mean to the dataset, with the individual aerofoils
  in blue.}
  \label{fig:airfoil_karcher_mean}
\end{figure}

Figure~\ref{fig:airfoil_registration_eg} shows a registration between two
aerofoils from the UIUC database \citep{seliguiuc}, normalised to unit chord
and resampled uniformly in arc length. The velocity field is defined on
$\mathcal{X} = [-0.10, 1.10] \times [-0.25, 0.25]$ and integrated with
$Q = 10$ Runge--Kutta steps. We use three stages with control lattices
$8 \times 8$, $16 \times 16$ and $32 \times 32$, boundary resolutions
$P = 100, 300, 500$, and regularisation weights
$\lambda = 5 \times 10^{-2}, 10^{-3}, 10^{-5}$. The full registration takes
around 170 optimiser iterations, reaching a loss below $10^{-6}$.

The same machinery can be used to compute a mean shape. Minimising the total
Chamfer distance to all $N$ examples directly over $P$ free points is a
high-dimensional problem, so we instead take the pointwise mean of the
examples and register it to the whole dataset, minimising
$\sum_{k=1}^N d_C(\{\varphi_1(x_R^i)\}_i, A_k) + \lambda \|v_c\|_{\mathrm{Sob}}^2$,
where $x_R^i = N^{-1} \sum_k x_k^i$ and $A_k$ is the $k$-th example.
Figure~\ref{fig:airfoil_karcher_mean} shows the result for 100 aerofoils.

\section{Experimental Details}
\label{app:experiments}

\paragraph{Datasets and solvers.} 2D aerofoils. We use the UIUC aerofoil database, excluding multi-element entries (names containing flap, slat, main-flap or 30p-30n) and files that fail basic checks on point count, chord length and thickness. This leaves 1657 aerofoils. Each is chord-normalised, resampled to 400 points with the trailing and leading edges pinned, and registered to the mean contour with a smooth velocity field (30×30 B-spline lattice, RK4 flow) to give pointwise correspondences. Aerodynamics are computed with NeuralFoil (medium model) at $\alpha=4^\circ$ and $Re=10^6$. The objective is $L/D \times$ analysis confidence, which down-weights shapes where the surrogate is outside its training regime. Invalid geometries (failed thickness or leading-edge checks) and failed or non-finite evaluations score zero. 3D wings. We use the 776 OptiWing3D wings, each a lift-constrained drag minimisation, stored as 9 spanwise sections of 192 points. We resample each section to TE/LE-pinned points and loft the span from 9 to 27 stations with PCHIP. No wings are discarded. All wings share one frame and registration lattice (36³). Wings are evaluated with AeroBuildup (medium NeuralFoil section polars plus analytical induced and wave drag models) at Mach 0.594, $Re=5.58\times10^{6}$ (dataset medians) and fixed $\alpha=4^\circ$. The objective is $L/D$ of the full wing. Wings failing topological checks (monotonic span, section thickness, leading-edge location, plausible chord) or returning non-finite aerodynamics score zero.

\paragraph{Bayesian optimisation loop.} Every method is run for $T=300$ evaluations, which is $n_0=50$ initial designs plus 250 BO iterations. For each seed, the initial designs are drawn without replacement from dataset designs lying inside the search box, so no extra evaluations are spent on random shapes. For wings, all spaces share the same 50 initial wings per seed. Each shape space is searched in a $\pm3\sigma$ box mapped to $[0,1]^d$. The default surrogate is a GP with an ARD Matérn-5/2 kernel (scale-kernel prefactor, standardised targets), refit every iteration, with analytic LogEI. LogEI is evaluated on a scrambled-Sobol candidate pool of 1024 points (2D) or 2048 (3D), and the maximiser is selected. In SWWL runs, the kernel acts on the decoded geometry. Each candidate shape is embedded with a sliced-Wasserstein Weisfeiler–Lehman embedding (2 WL iterations, 32 projections, 64 quantiles, 200 arc points per aerofoil or a 9×64 ring/section graph per wing), followed by an RBF kernel with a learned bandwidth, 300 training steps and refitting every second evaluation, over a pool of 512 candidates. This makes the surrogate independent of the parametrisation. All runs use a single TuRBO trust region centred on the incumbent, with initial side length 0.8 (maximum 1.6, minimum $2^{-7}$) and per-dimension scaling by the ARD lengthscales (normalised to unit geometric mean), clipped to the search box. The length doubles after 3 consecutive successes (an improvement of at least $10^{-6}$) and halves after 5 consecutive failures in 2D ($\max(4,d)=15$ in 3D). If it falls below $2^{-7}$, it resets to 0.8 around the current incumbent. The Sobol baseline uses scrambled Sobol sequences over the same CST box.

\section{Sensitivity to Box Size}
\label{app:boxsize}

\begin{figure}[htbp]
    \centering
    \includegraphics[width=0.9\textwidth]{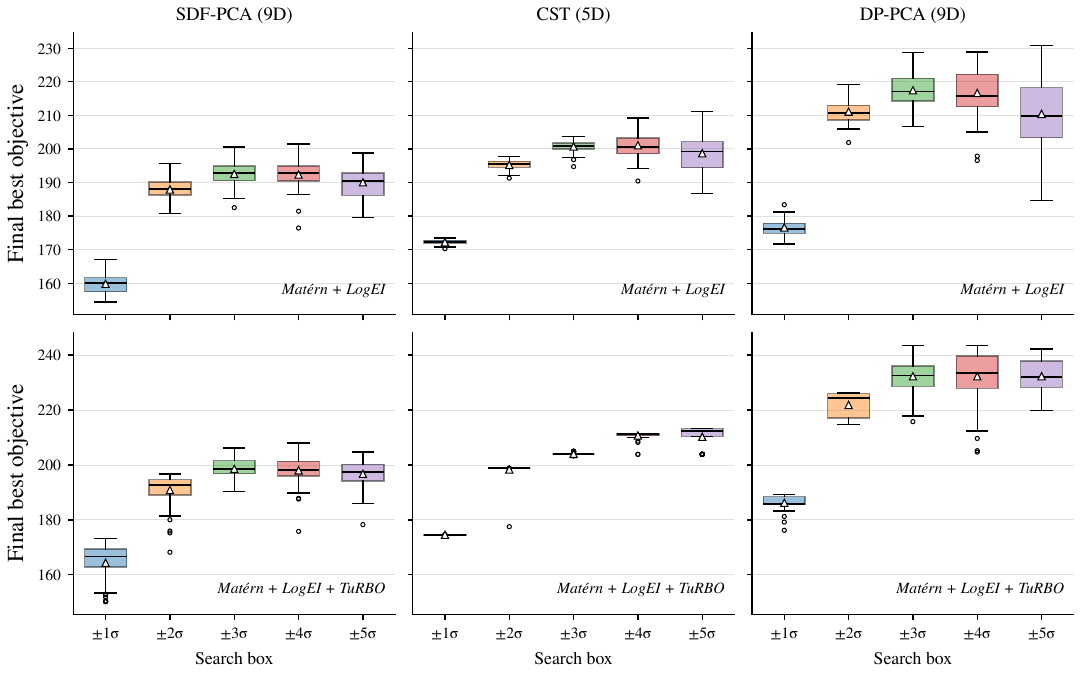}
    \caption{Final best objective after $T=300$ evaluations as the search box grows from $\pm1\sigma$ to $\pm5\sigma$. Widening the box from $\pm1\sigma$ to $\pm3\sigma$ markedly improves the final objective in every space and with both optimisers, since larger ranges admit more extreme shapes that lie outside the bulk of the dataset. Beyond $\pm3\sigma$ the gains saturate: the PCA spaces show no further improvement, with a slightly lower mean and larger spread at $\pm5\sigma$ as the search volume grows, whereas CST with TuRBO continues to improve up to $\pm5\sigma$.}

    \label{fig:appendix_sigma_box}
\end{figure}

\begin{figure}[htbp]
    \centering

    \begin{subfigure}[b]{0.48\textwidth}
        \centering
        \includegraphics[width=\textwidth]{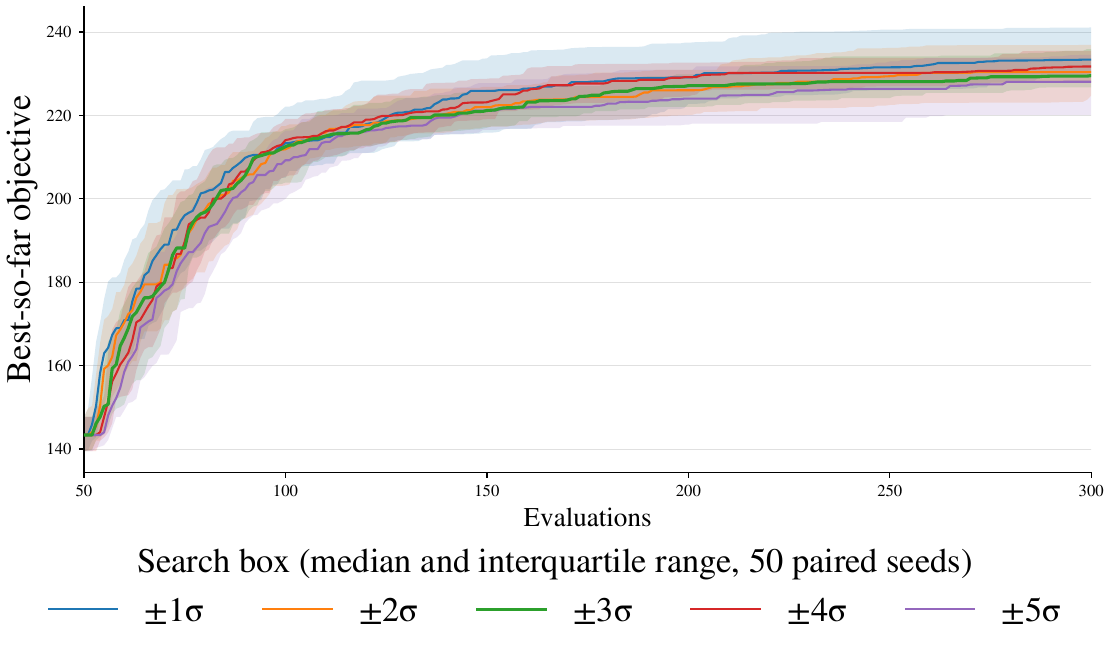}
        \caption{Performance of unbounded TurBO}
        \label{fig:appendix_sigma_unbounded}
    \end{subfigure}
    \hfill
    \begin{subfigure}[b]{0.48\textwidth}
        \centering
        \includegraphics[width=\textwidth]{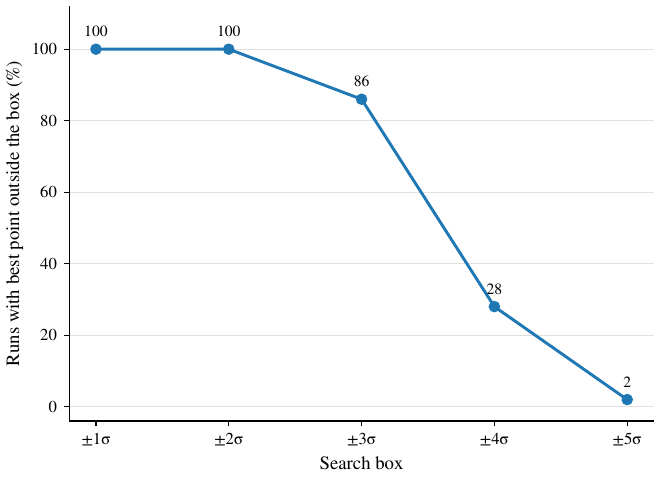}
        \caption{Best solutions are in distant regions of the search space}
        \label{fig:appendix_sigma_out}
    \end{subfigure}

    \caption{Left: Best-so-far objective against the number of evaluations for Mat\'ern\,+\,LogEI with an
  unbounded (may leave the box) TuRBO trust region on 9D DP-PCA, for search boxes of half-width $\pm1\sigma$ to
  $\pm5\sigma$. Lines show the median over 50 seeds and shaded bands the interquartile range;
  each run uses 50 shared initial aerofoils and 300 evaluations in total. As expected, unbounded TuRBO is largely agnostic to the box size as its trust region is not
  clipped to the box, so the box serves mainly to set the coordinate scale and starting domain
  and does not limit where the search can go. Right: Share of runs whose best solution lies outside the search box, as a function of
  the box half-width.
  Runs use Mat\'ern\,+\,LogEI with an unbounded TuRBO trust region on 9D DP-PCA, with 50 seeds,
  50 initial and 300 total evaluations per run. We see that a tight box excludes better
  solutions and constrains the search. }
    \label{fig:appendix_both}
\end{figure}

\begin{figure}[htbp]
    \centering
    \includegraphics[width=0.9\textwidth]{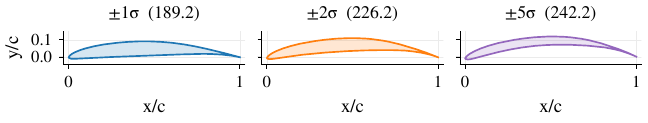}
  \caption{Best aerofoils found by Mat\'ern+TuRBO under search boxes of $\pm1\sigma$, $\pm2\sigma$ and $\pm5\sigma$. The best objective value is given in parentheses above each panel. Widening the box improves the best objective ($189.2 \to 226.2 \to 242.2$) - the winning shapes are far from the typical atlas aerofoil. As in the earlier ablations, the best foils once again require quite extreme, large coordinates in the latent space, often far from the centre of the box, so the achievable objective depends strongly on how large a region we allow.}
    \label{fig:appendix_airfoils_w_sigma}
\end{figure}

\paragraph{Effect of the search range.}
In this section we explore the impact of changing the size of our search space when using bounded methods to find optimal aerofoils. Figure~\ref{fig:appendix_sigma_box} reports the final best objective as the box half-width grows from $\pm1\sigma$ to $\pm5\sigma$, and shows that a wider box lets BO reach better optima up to about $\pm3\sigma$, after which performance saturates. Figure~\ref{fig:appendix_both} then examines TuRBO with an unbounded trust region: panel~(\subref{fig:appendix_sigma_unbounded}) shows best-so-far curves for DP-PCA, which are largely insensitive to the box size, and panel~(\subref{fig:appendix_sigma_out}) shows how often the best solution lies outside the box, indicating how much a tight box constrains the search. Finally, Figure~\ref{fig:appendix_airfoils_w_sigma} compares the best aerofoils found under $\pm1\sigma$, $\pm2\sigma$ and $\pm5\sigma$, illustrating that the higher objective values come from shapes far from the typical dataset aerofoil.


\newpage
\section{Sensitivity to number of PCA components}
\label{app:PCA}
\begin{figure}[htbp]
    \centering
    \includegraphics[width=0.9\textwidth]{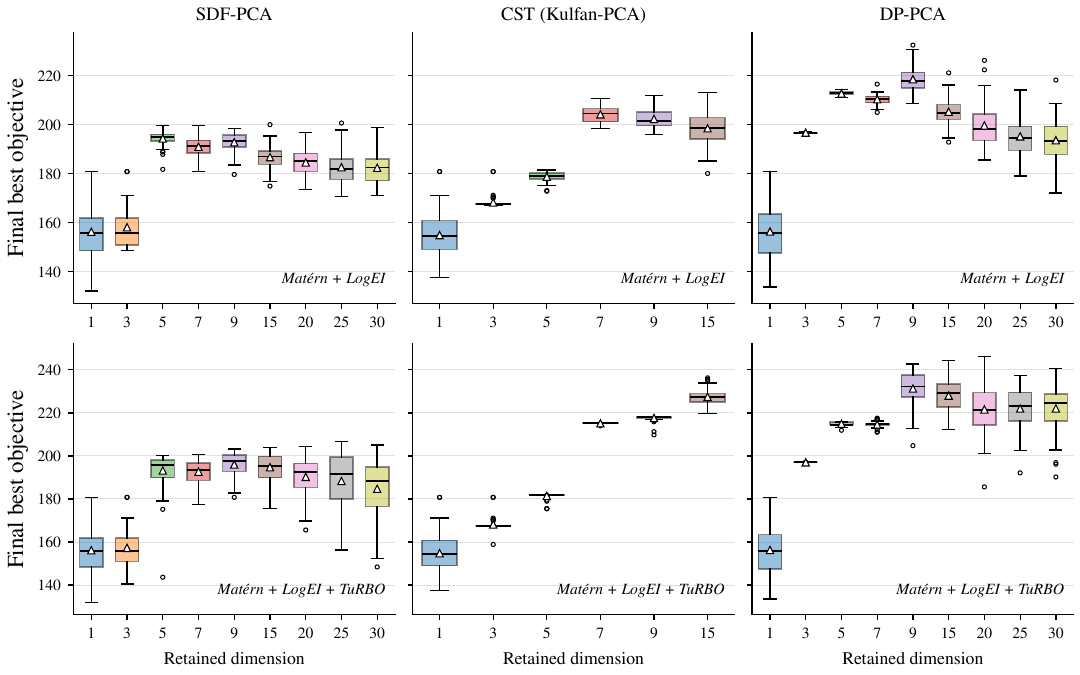}
  \caption{Final best objective after $T=300$ evaluations as a function of the number of retained dimensions, all in a $\pm3\sigma$ box. Boxes show the median and interquartile range over 50 seeds, whiskers the range of non-outlying values, triangles the mean and circles outliers. Very low-dimensional spaces (1--3 modes) cannot represent good aerofoils and perform worst in every space. Performance then rises sharply and peaks at intermediate dimensions (5--9 modes for SDF-PCA and DP-PCA), after which it declines slowly as the larger search space becomes harder to optimise, while DP-PCA remains the strongest space at its optimum. CST-PCA gains up to 7--9 dimensions and, with TuRBO, continues to improve up to 15. TuRBO improves the final objective in most settings, most visibly at higher dimensions, where it limits the loss of performance seen with global Mat\'ern\,+\,LogEI.}

    \label{fig:appendix_dim_box}
\end{figure}

In this section we look at how the number of retained dimensions affects optimisation in each shape space. Figure~\ref{fig:appendix_dim_box} shows the final best objective as the dimension varies, with performance peaking at an intermediate dimension (around 9 for DP-PCA) rather than increasing with dimension.

\section{RF cavity details}

In this section we include supplementary details and context to the RF cavity application.

In table \ref{tab:eic_design_spec}, 
we detail an external design specification \citep{EIC_report} for our RF cavities.

\begin{table}[h!]
    \centering
    \begin{tabular}{cc}
        \toprule
        \textbf{Parameter} & \textbf{Value/Unit} \\
        \midrule
        Frequency $f_0$ & \SI{394}{\mega\hertz}
        \\
        Transverse voltage $V_t$ & \SI{3.5}{\mega\volt} \\
        $E_p/E_t$ & 4.895 \\
        $B_p/E_t$ & \SI{8.702}{\milli\tesla\per(\mega\volt\per\metre)} \\
        Aperture diameter & \SI{100}{\milli\metre} \\
        \bottomrule
    \end{tabular}
    \caption{Design specification for the 394 MHz RFD cavity \citep{EIC_report}}
    \label{tab:eic_design_spec}
\end{table}

\begin{figure*}[t]
\centering
\includegraphics[width=\textwidth]{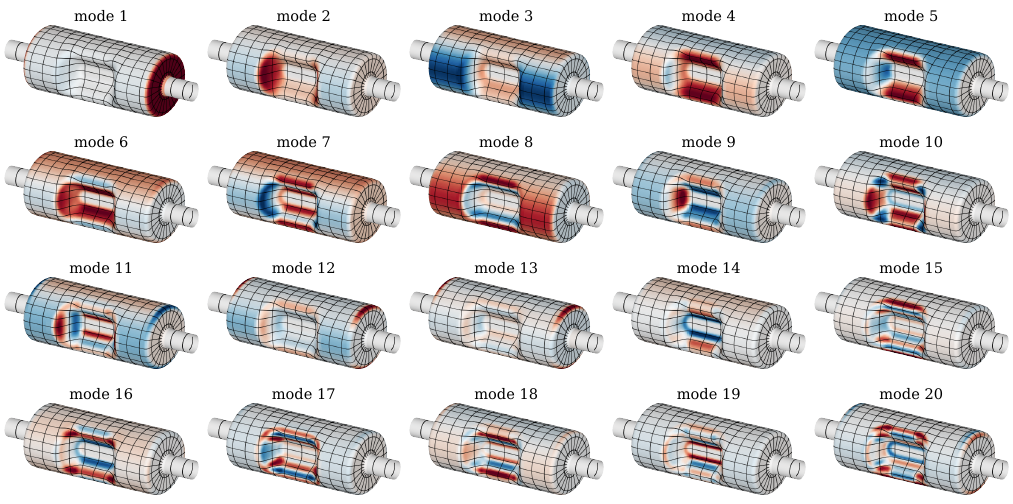}
\caption{The 20 shape modes of the cavity dataset, each drawn as the wall displacement it produces on the mean design per one dataset standard deviation, normal to the wall (red outwards, blue inwards; each panel on its own colour scale), with the correspondence grid.}
\label{fig:shape-modes}
\end{figure*}

In Figure \ref{fig:shape-modes}, we plot the first 20 eigenmodes arising from the principal component analysis of our intial dataset. Here, only the normal component of their displacement field is plotted. We see that the leading modes account for interpretable, bulk motion of the mean cavity while the trailing modes refine increasing subtle features.

\begin{figure*}[t]
\centering
\includegraphics[width=\textwidth]{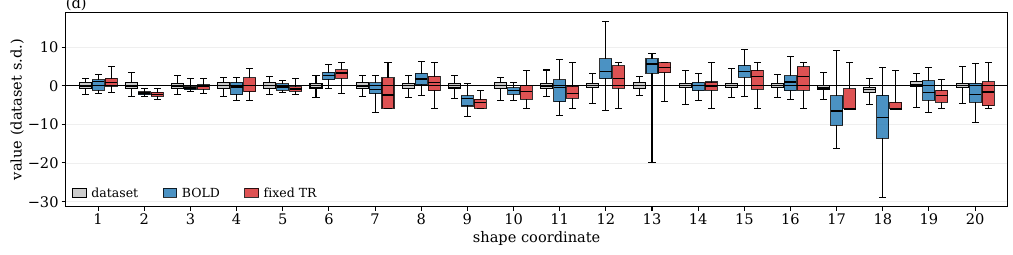}
\caption{Shape coordinates of the dataset and of the pooled Pareto fronts of BOLD and the fixed-TR ablation.}
\label{fig:box-and-whisker}
\end{figure*}

In Figure \ref{fig:box-and-whisker}, we make a box-and-whisker plot of 1) our initial dataset, 2) our BOLD multi-objective pooled Pareto front and 3) the multi-objective pooled Pareto front from our fixed bounds ablation in Section \ref{sec:eg-cavities}, all in the above shape coordinates. We can clearly see that both fronts draw from a significantly larger coordinate space. The ablation remains capped at $6\sigma$, while BOLD's ranges are even more extreme -- exceeding $25 \sigma$ for mode 18.

\begin{figure*}[h]
\centering
\includegraphics[width=\textwidth]{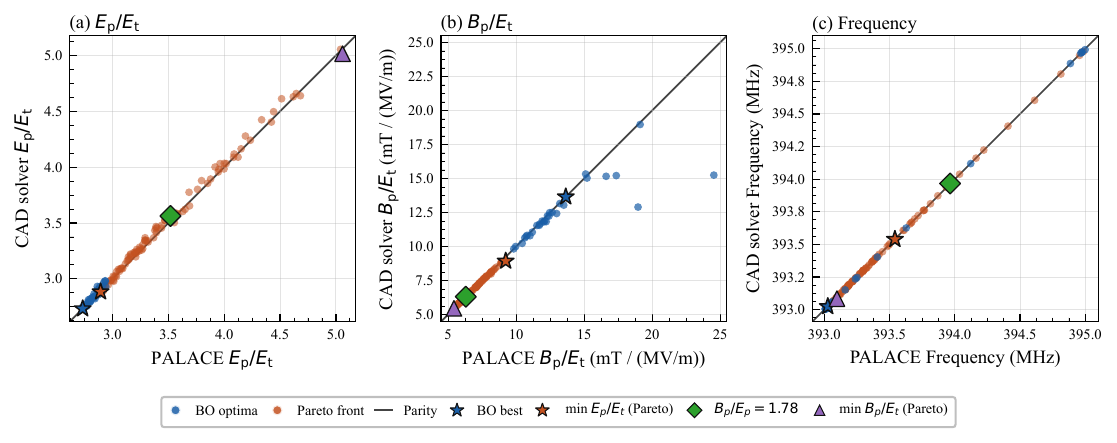}
\caption{Comparison of PALACE finite-element solver with CAD solver (CST Microwave Studio) for a total of 130 solves. Optima from the single objective optimisation runs minimising $E_p/E_t$ are shown in blue while Pareto-optimal solutions from the multiobjective runs are shown in orange.}
\label{fig:validate}
\end{figure*} 

To confirm the fidelity of the open-source PALACE solver, designs were passed to the CAD solver (CST Microwave Studio) for evaluation. See Fig. \ref{fig:validate}. Agreement is excellent for $E_p/E_t$ and Frequency, with mean absolute percentage errors below 3\,\%. The discrepancy at larger $B_p/E_t$ (Fig. \ref{fig:validate}(b)) is attributed to small differences in the mesh between the two solvers. Geometries with high $B_p/E_t$ exhibit tight curvatures which require a very fine mesh to resolve such features. Importantly, these designs do not yield favourable objective values and are therefore excluded from subsequent analysis. The conclusions drawn from the optimisations results are unaffected by this local discrepancy.

\end{document}